\documentclass{article}

\usepackage{arxiv}

\usepackage[utf8]{inputenc} 
\usepackage[T1]{fontenc}    
\usepackage{hyperref}       
\usepackage{url}            
\usepackage{booktabs}       
\usepackage{amsfonts}       
\usepackage{nicefrac}       
\usepackage{microtype}      
\usepackage{lipsum}
\usepackage{natbib}
\usepackage{float}
\usepackage{lipsum}
\usepackage{graphicx}
\usepackage{amsmath}
\usepackage{caption}
\usepackage{subcaption}
\usepackage{amsfonts}
\usepackage{float}
\usepackage{caption} 
\title{Conformal Prediction Intervals for Neural Networks Using Cross Validation}

\author{
  Saeed Khaki \\
  Department of Statistics\\
  Iowa State University\\
  Ames, IA, 50011 \\
  \texttt{skhaki@iastate.edu} \\
   \And
 Dan Nettleton \\
  Department of Statistics\\
  Iowa State University\\
  Ames, IA, 50011 \\
  \texttt{dnett@iastate.edu } \\
}

\begin{document}
\maketitle

\begin{abstract}

Neural networks are among the most powerful nonlinear models used to address supervised learning problems. Similar to most machine learning algorithms, neural networks produce point predictions and do not provide any prediction interval which includes an unobserved response value with a specified probability. In this paper, we proposed the $k$-fold prediction interval method to construct prediction intervals for neural networks based on $k$-fold cross validation.  Simulation studies and analysis of 10 real datasets are used to compare the finite-sample properties of the prediction intervals produced by the proposed method and the split conformal (SC) method. The results suggest that the proposed method tends to produce narrower prediction intervals compared to the SC method while maintaining the same coverage probability. Our experimental results also reveal that the proposed $k$-fold prediction interval method produces effective prediction intervals and is especially advantageous relative to competing approaches when the number of training observations is limited.
\end{abstract}




\section{Introduction}

Neural networks are mathematical functions that map some set of input values to output values \citep{Goodfellow2016}. Neural network models belong to the class of representation learning methods that automatically discover the underlying representations of data. A neural network model is composed of multiple processing layers that each transforms the representation at one level into a more abstract representation starting from the raw input \citep{LeCun2015}. As such, a very complex function can be learned if we combine enough transformations. Such transformations are obtained by stacking nonlinear modules \citep{LeCun2015,Goodfellow2016}. Neural networks are also known to be universal approximators which means that regardless of the function we want to learn, a large enough neural networks can represent such a function \citep{hornik1989multilayer}. However, learning the desired
function using neural networks is challenging and there is no guarantee that we can find the right parameters for the neural networks \citep{Goodfellow2016}. 

Similar to most machine learning methods for prediction, neural networks usually produce point predictions without any information about how far point predictions are from the ground truth response variables. Because point predictions produced by neural networks do not assess the prediction error from the same data used to generate point predictions, neural networks are lacking in inferential capability from a statistical standpoint. In this paper, we develop prediction intervals based on neural network point predictions that produce a range of values including an unknown continuous univariate response with any specified level of confidence.

Following the setup in \citep{zhang2019random}, $(X,Y) \in \mathbb{R}^p \times \mathbb{R} $ denote the predictor-response pair randomly sampled from some distribution $\mathbb{G}$, where $p$ is the number of predictors and $Y$ is a continuous univariate response. We develop a prediction interval for the observation $(X,Y)$ denoted as $I_\alpha(X,C_n)$ that will cover the true response variable with the probability $1-\alpha$, where $C_n$ is a training set including observations $(X_1,Y_1),...,(X_n,Y_n) \overset{\text{iid}}{\sim} \mathbb{G} $ and $(X,Y)$ is independent of the training set $C_n$.

Several approaches have been proposed to construct prediction intervals for neural networks. For example, \cite{hwang1997prediction} proposed an asymptotic approach to construct prediction intervals for neural networks. They estimated the asymptotic variance of the neural network predictions and used the $1-\alpha/2$ quantile of a $t$-distribution to create prediction intervals. \cite{de1998prediction} constructed prediction intervals for neural networks based on the asymptotic variance of the estimated parameters of the neural networks. \cite{khosravi2010lower} proposed a lower upper bound estimation method to construct two outputs for a neural network model  for estimating the prediction interval bounds. \cite{kivaranovic2019adaptive} proposed a distribution-free split conformal prediction interval for neural networks. They designed a prediction interval network which had three outputs to estimate the median and the lower and upper bounds of prediction intervals.

The conformal prediction interval framework is a general method to construct prediction intervals and provides distribution-free predictive inference \citep{vovk2005algorithmic}. Many prediction interval methods have been proposed based on conformal inference. For example, \cite{lei2018distribution} proposed a distribution-free predictive inference for regression leading to split conformal (SC) prediction intervals. However, the SC method may not always have a good performance, especially when the sample size is small because the SC approach uses only half of the data to train the regression function, which may not be sufficient. In this creative component, we propose a $k$-fold prediction interval method to construct prediction intervals based on $k$-fold cross validation. This method tends to produce narrower prediction intervals compared to SC intervals while maintaining the same coverage probability. Our experimental results suggested that the proposed $k$-fold prediction interval method is effective and especially advantageous when the number of training observations is limited.

The remainder of this paper is organized as follows. Section 2 describes the methodology. Section 3 presents the simulation study. Section 4 explains the data analysis results. Finally, we conclude the paper in section 5.



\section{$k$-fold Conformal Prediction Intervals}

The conformal prediction interval framework is a general approach for efficiently constructing prediction intervals \citep{vovk2005algorithmic}. To decrease the computational cost of the full conformal method, Lei et al. proposed split conformal prediction intervals, which are considerably computationally more efficient than the full conformal method \citep{lei2018distribution}. The SC prediction interval algorithm includes the following steps:
\begin{enumerate}

\item Randomly split $\{1,...,n\}$ into two equal-sized subsets $L1$ and $L2$.

\item Train a regression function from $\{(X_i,Y_i): i \in L_1 \}$ to estimate the mean function denoted as $\hat{m}_{n/2}(X)$.

\item For $i \in L_2$, compute the prediction error $D_i=Y_i-\hat{Y_i}$, where $\hat{Y_i}=\hat{m}_{n/2}(X_i)$.

\item Construct the prediction interval with coverage probability $1-\alpha$ for $Y$ as $[\hat{Y}-D_{[n/2,\alpha/2]}, \hat{Y}+D_{[n/2,\alpha/2]}]$, where $D_{[n/2,\eta]}$ is the $\eta$ quantile of the empirical distribution of $D_1,..., D_{n/2}$.

\end{enumerate}

Although SC method generates reliable prediction intervals, it may not always have good performance, especially when the sample size is small. The SC method uses half of the data to train the regression function which may not always be sufficient. All observations do not get a chance to contribute to the construction of empirical distribution of errors. In this paper, we propose a new method to construct prediction intervals based on $k$-fold cross validation called $k$-fold conformal prediction interval. The $k$-fold conformal prediction interval algorithm includes the following steps:

\begin{enumerate}
\item Randomly split $\{1,...,n\}$ into $k$ equal-sized subsets denoted as $L_1$, $L_2$,..., $L_k$.

\item For each $L_j$ where $j\in\{1,...,k\}$ do the following:
\begin{enumerate}
\item Train a regression function from $\{(X_i,Y_i): i \in \bigcup\limits_{r=1 }^{k} L_{r}, \; r\neq j \}$ to estimate the mean function denoted as $\hat{m}_{j}(X)$.
\item For $i \in L_j$, compute the prediction error $D_i=Y_i-\hat{Y_i}$, where $\hat{Y_i}=\hat{m}_{j}(X_i)$.

\end{enumerate}

\item Construct the prediction interval with coverage probability $1-\alpha$ for $Y$ as $[\hat{Y}-D_{[n,\alpha/2]}, \hat{Y}+D_{[n,\alpha/2]}]$, where $D_{[n,\eta]}$ is the $\eta$ quantile of the empirical distribution of $D_1,..., D_{n}$.

\end{enumerate}

The proposed $k$-fold conformal prediction interval requires estimation of $k$ regression functions which results in the empirical distribution of errors based on the all training data observations. Thus, the proposed method's computational cost is on the order of $k$ times that of the SC method. Each training set is larger too, so there could be added expense.

\section{Simulation Study}

To evaluate the finite-sample performance of the proposed approach, we conducted a simulation study to compare our proposed method to the SC method with respect to coverage rate and interval width performance measures. Data are simulated from an additive error model: $Y=m(X)+\epsilon$, where $X=(X_1,...,X_p)$ with $p=10$ and $X \sim \mathcal{N}(0,\varSigma_p)$, where $\varSigma_p$ is an AR(1) covariance matrix with $\rho=0.6$ and diagonal values equal to 1. We considered three factors, namely the distribution of the error terms, the choice of mean function $m()$, and the number of training observations $n$. Following \citep{zhang2019random}, we considered the following factorial design for these three factors:\\

\begin{itemize}
\item \textbf{Mean functions}: 
\begin{enumerate}
\item linear: $m_1(x)=x_1+x_2$
\item nonlinear: $m_2(x)=2\exp(-|x_1|-|x_2|))$
\item nonlinear with interaction: $m_3(x)=2\exp(-|x_1|-|x_2|))+x_1x_2$
\end{enumerate}

\item \textbf{Distributions of errors}: 

\begin{enumerate}
\item homoscedastic: $\epsilon \sim \mathcal{N}(0,1)$
\item heavy-tailed: $\epsilon \sim \frac{t_3}{\sqrt{3}}$, where $t_3$ is a $t$-distribution with 3 degrees of freedom.
\item heteroscedastic: $\epsilon \sim \mathcal{N}(0,\frac{1}{2}+\frac{1}{2} \frac{|m(X)|}{E|m(X)|})$
\end{enumerate}




\item \textbf{Training Sample sizes}: $n=500, 2500$, and $5000$ 

\end{itemize}


The full-factorial design has 27 different simulation scenarios. The following hyperparameter were used to train the neural network model. The neural network model has 2 fully connected layers with 15 neurons in each layer. We investigated different activation functions, such as ReLU and tanh, and found that ReLU had the best overall performance. Only results for ReLU are reported here. All weights were initialized with the Xavier method \citep{Glorot2010}. We used stochastic gradient descent (SGD) with a mini-batch size of 32. The Adam optimizer \citep{Kingma2014} with learning rate of 0.03\% was used to minimize the loss function. The model was trained for 20,000 iterations.\\



\subsection{Evaluation of Coverage Rates and Interval Widths}

\begin{figure}[H]
\centering
\includegraphics[scale=0.38]{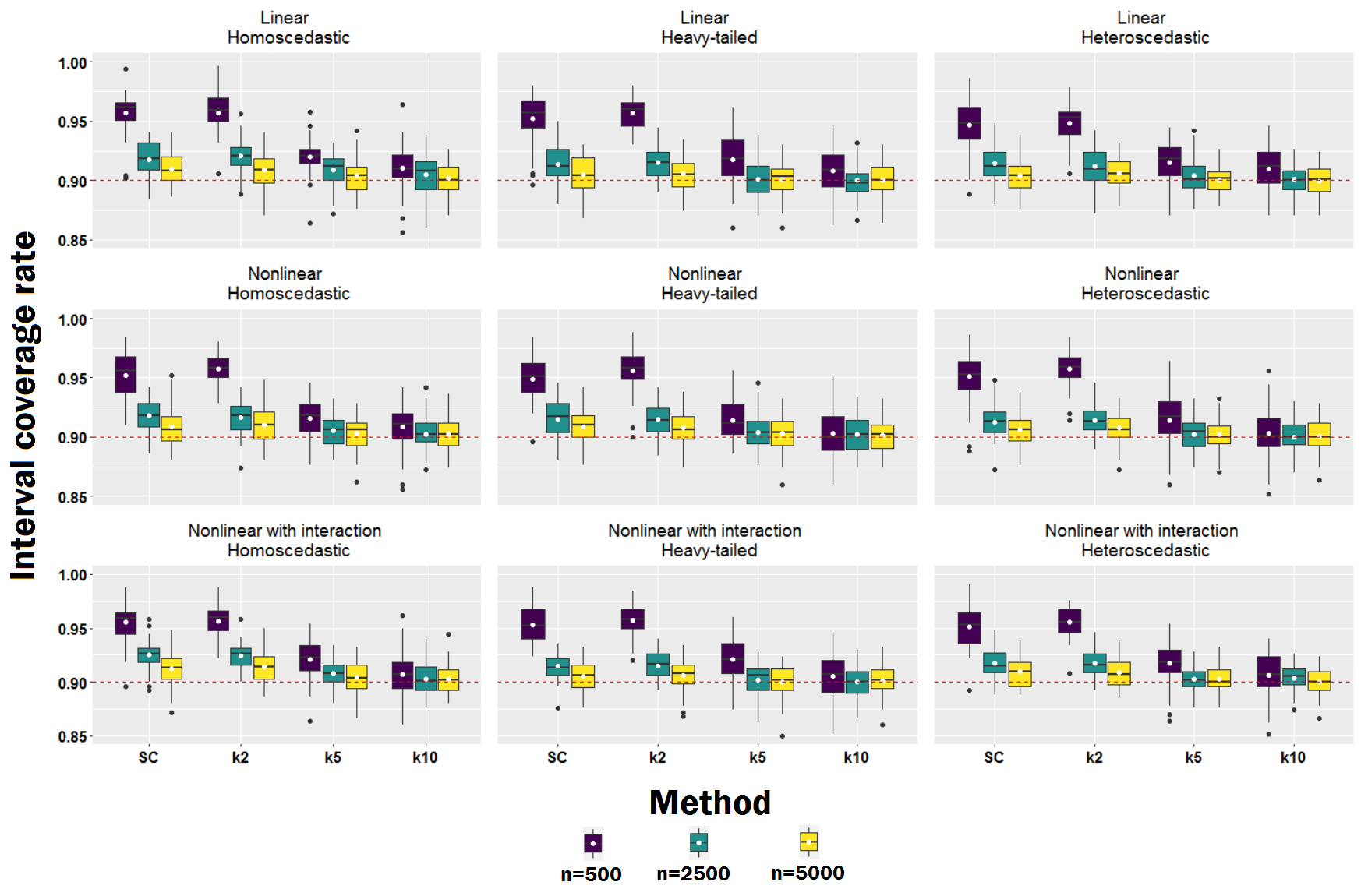}
\caption{Boxplots of the coverage rate estimates of the split conformal method (SC), 2-fold prediction interval method (k2), 5-fold prediction interval method (k5), and 10-fold prediction interval method (k10). The white circle in each boxplot is the average of the 50 coverage rate estimates for each simulation scenarios. The dashed red lines show the nominal coverage level which is set to be 0.9 in our study.}\label{fig:coverage}
\end{figure}

The nominal coverage level was set at 0.9 for the construction of all prediction intervals in the simulation study. 50 datasets were simulated for each simulation scenario. We also generated 500 test samples independently from the joint distribution of $(X,Y)$ for each simulation scenario. We defined the coverage rate as the percentage of response values contained in their corresponding prediction intervals for the test data. We estimated the coverage rate by mean of coverage rates obtained from 50 simulated datasets for each simulation scenario. To evaluate the effect of $k$ on the performance of prediction intervals, we considered three different $k$ values, namely 2, 5 and 10. Figure \ref{fig:coverage} compares the coverage rate estimates of the SC method, 2-fold prediction interval method, 5-fold prediction interval method, and 10-fold prediction interval method. The white circle in each boxplot is the average of the 50 coverage rate estimates for each simulation scenario. The estimates of the coverage rates for k5 and k10 are closer to 0.9 (the nominal level) especially for the larger sample sizes compared to the SC and k2 methods. The SC and k2 methods tend to over-cover the response values based on coverage rates especially when the sample is small because these methods use only half of the training data to train the mean function which may be insufficient when $n$ is small. Thus, the prediction errors would be larger resulting in a wider prediction intervals and over-coverage. The SC method uses only half of the training data to find the prediction errors for the other half of the training data which makes it similar to the k2 method except that k2 method finds prediction errors for all training data. As such, SC and k2 methods have very similar boxplots for the coverage rates. As the sample size increases, the coverage estimates of all methods become more concentrated around the nominal level due to having adequate information to estimate the mean functions using neural networks. The results suggest that all competing methods showed stable behavior in terms of the coverage rate estimates across all factors including the mean functions and the choice of error distributions.

\begin{figure}[H]
\centering
\includegraphics[scale=0.38]{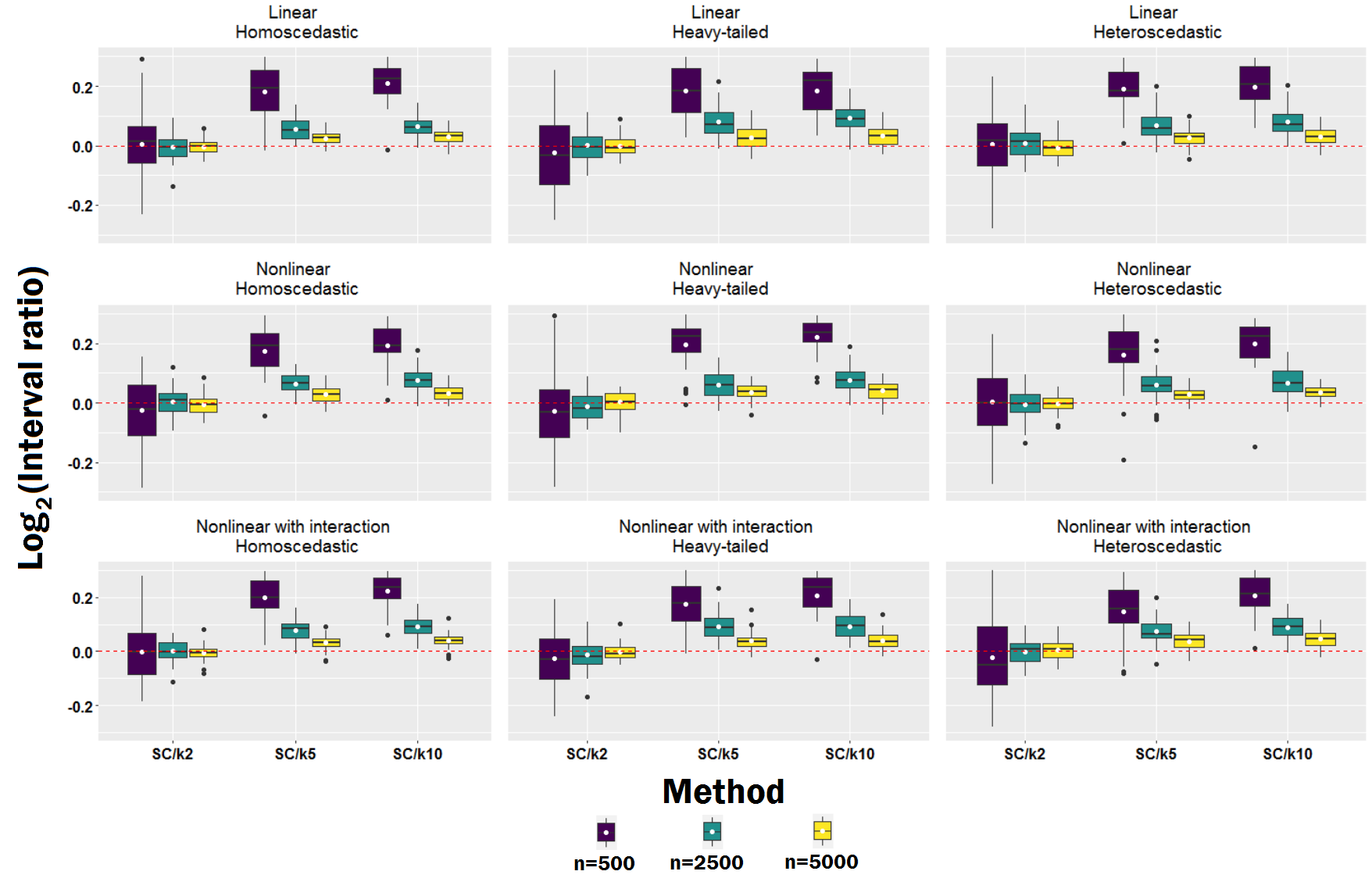}
\caption{Boxplots of the $log_2$ ratios of split conformal (SC) interval widths to 2-fold prediction interval (k2) widths, 5-fold prediction interval (k5) widths, and 10-fold prediction interval (k10) widths. The white circle in each boxplot is the average of the 50 $log_2$ interval width ratios for each simulation scenario.}\label{fig:width}
\end{figure}

To evaluate the prediction interval widths, we averaged the 500 test cases' interval widths for each simulated dataset. To better compare the proposed $k$-fold prediction interval method with the SC method, we computed the ratio of the SC interval width to the $k$-fold prediction interval width. Figure \ref{fig:width} shows the $log_2$ ratios of the interval widths. As shown in Figure \ref{fig:width}, interval width decreases as the sample size increases due to availability of enough training data to estimate the mean functions well. The SC and k2 produce intervals of approximately the same width as indicated by log ratios close to zero. However, the SC method tends to have a slightly smaller interval width, especially when the sample size is small. Results demonstrate that k5 and k10 prediction intervals are smaller than intervals constructed by the SC method. The k5 and k10 prediction interval methods have a comparable performance in terms of prediction interval widths, which indicates that increasing $k$ in the $k$-fold prediction interval method does not always improve the performance. This indicates an opportunity to choose $k$ to obtain narrow prediction intervals while maintaining low computational costs. The log ratios show that the k5 and k10 methods have the biggest advantages over SC intervals, in terms of width, when the training sample size is small.

\section{Data Analysis}

To evaluate the performance of our proposed prediction interval method on real-world datasets, we selected 10 datasets from UC Irvine Machine Learning Repository website which are summarized in the Table \ref{tab:summary}.

\begin{table}[H]
\centering
\begin{tabular}{|c|c|c|c|}
\hline
No.& Name of dataset & Number of predictors & Number of observations\\
\hline
1&Power Plant&4&9,568\\
\hline
2&Facebook Metrics & 18& 500\\
\hline
3& Parkinsons Telemonitoring& 21& 5,875\\
\hline
4& Bodyfat& 13& 252\\
\hline
5& Residential Building& 106 & 372\\
\hline
6 & Real Estate Valuation & 5& 414\\
\hline
7& Wine Quality & 11 & 4898\\
\hline
8& Aquatic Toxicity &8&546\\
\hline
9& Fish Toxicity & 6 &908\\
\hline
10&Energy Efficiency &8 & 768\\
\hline
\end{tabular}\caption{The summary of real datasets.}\label{tab:summary}
\end{table}

To obtain the data analysis results in this section, we used the following hyperparameters for neural networks. The neural networks model has 2 fully connected layers with 10 neurons in each layer. As in the simulation described in the section 3, ReLU activation functions were used and all weights were initialized with Xavier method \citep{Glorot2010}. We used stochastic gradient descent (SGD) with a mini-batch size of 16. The Adam optimizer \citep{Kingma2014} with learning rate of 0.03\% was used to minimize the loss function. The model was trained for 25,000 iterations.\\

\begin{figure}[H]
\centering
\includegraphics[scale=0.32]{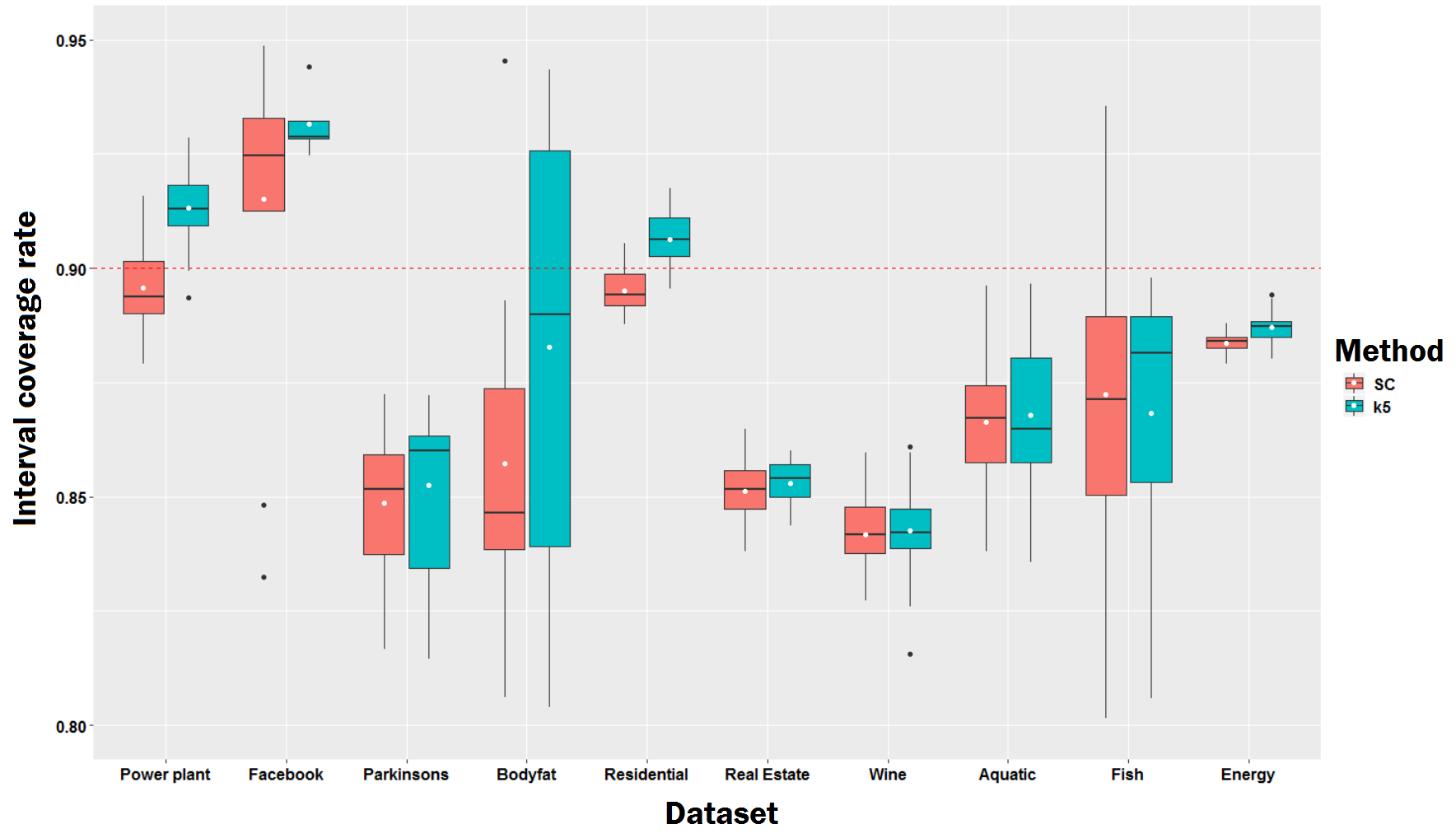}
\caption{Boxplots of the coverage rate estimates of split conformal method (SC) and 5-fold prediction interval method (k5). The white circle in each boxplot is the average of the 20 coverage rate estimates resulting from the cross validation procedure. The dashed red lines show the nominal coverage level which is set to be 0.9 in our study.}\label{fig:coverage_real}
\end{figure}

To estimate the coverage rates and interval widths, we used 5-fold cross validation which was repeated 20 times for each dataset. Since the simulation study suggested that the SC method tends to perform better compared to the 2-fold prediction interval method (k2), we did not use the k2 method in this section. We employed the 5-fold prediction interval method (k5) rather than the 10-fold prediction interval method (k10) to decrease the computational cost. Figure \ref{fig:coverage_real} compares the coverage rate estimates of the SC method and the k5 method. The white circle in each boxplot is the average of the 20 coverage estimates that resulted from the cross validation procedure.

\begin{figure}[H]
\centering
\includegraphics[scale=0.32]{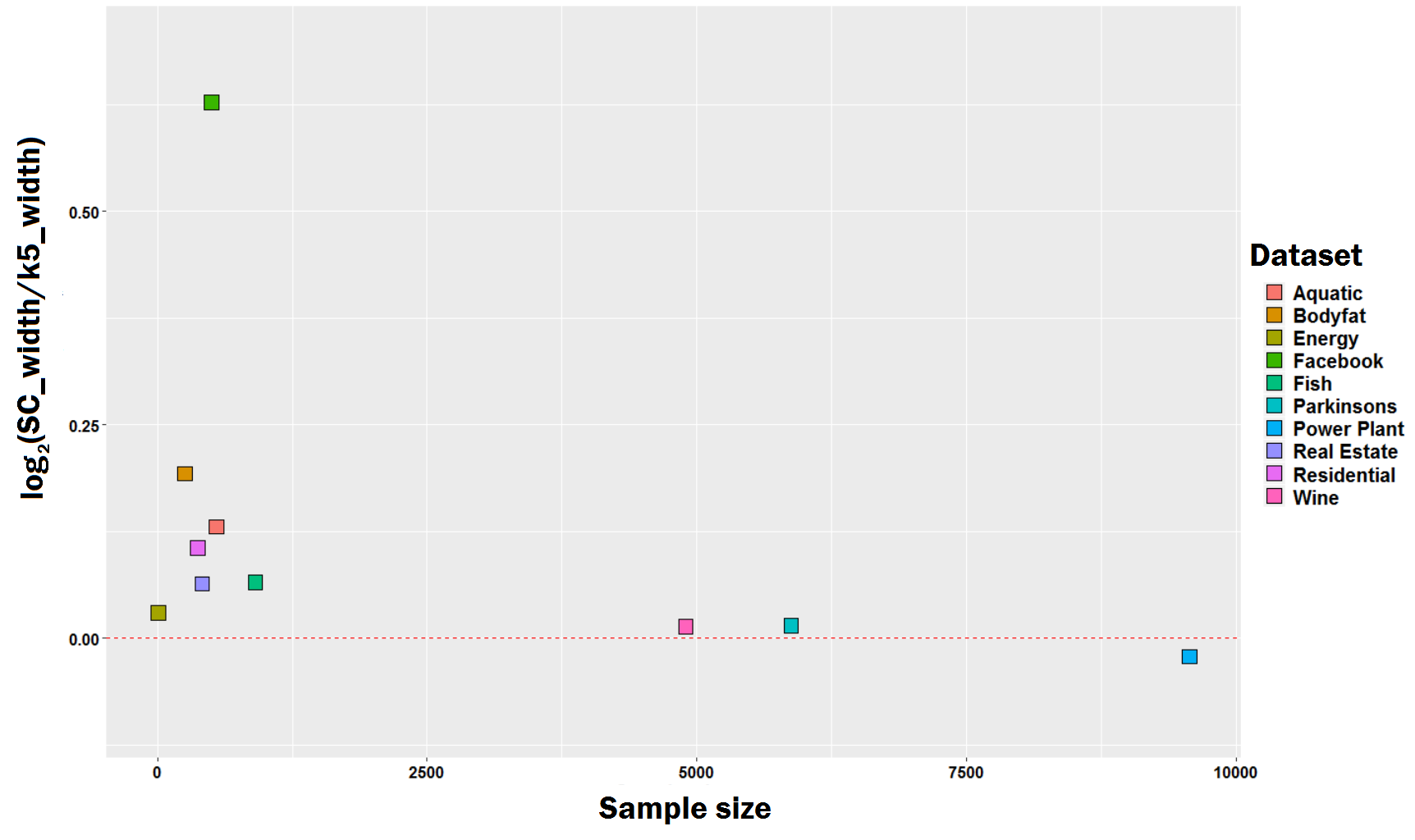}
\caption{Scatter plot of the $log_2$ ratios of the average split conformal (SC) interval widths to the average 5-fold prediction interval (k5) widths.}\label{fig:width_real}
\end{figure}

As shown in Figure \ref{fig:coverage_real}, the k5 prediction interval has higher average coverage than the SC method except for the Fish dataset. The results also indicate under coverage for some of the datasets. To evaluate the prediction interval widths, we averaged the 20 interval width estimates for each dataset. To better compare the proposed 5-fold prediction interval method with the SC method, we computed the ratio of the SC interval width to the width of the 5-fold prediction interval method. Figure \ref{fig:width_real} shows the $log_2$ ratios of the interval widths. The results indicate that the k5 method tends to have a smaller interval width than the SC method especially for the datasets with the smaller sample sizes since the $log_2$ ratios of the average SC interval widths to the average k5 widths are highest when the sample sizes are small. The results also reveal that the k5 method and SC method have a comparable performance for datasets with sufficiently large sample size. The SC method had a smaller interval width compared to the k5 method for the Power Plant dataset, but this was due to under coverage.

\section{Conclusion}

In this paper, we presented a conformal prediction interval method for neural networks using cross validation. The proposed method uses $k$-fold cross validation on the training data to estimate the empirical error distribution of the errors. Then, the quantiles of the empirical error distribution are used to construct prediction intervals for test data. Our experimental results indicate that the $k$-fold prediction interval method can efficiently construct prediction intervals for neural networks that compare favorably with intervals produced by the split conformal method. Our results suggest that the proposed method is well-suited for datasets with a small number of observations. We also found that the performance of the proposed model is somewhat sensitive to the choice of $k$. As such, it is important to tune $k$ in the proposed method to get narrow prediction intervals while maintaining low computational cost.

\bibliographystyle{plainnat}
\bibliography{references}  

\begin{thebibliography}{12}
\providecommand{\natexlab}[1]{#1}
\providecommand{\url}[1]{\texttt{#1}}
\expandafter\ifx\csname urlstyle\endcsname\relax
  \providecommand{\doi}[1]{doi: #1}\else
  \providecommand{\doi}{doi: \begingroup \urlstyle{rm}\Url}\fi

\bibitem[De~Vleaux et~al.(1998)De~Vleaux, Schumi, Schweinsberg, and
  Ungar]{de1998prediction}
Richard~D De~Vleaux, Jennifer Schumi, Jason Schweinsberg, and Lyle~H Ungar.
\newblock Prediction intervals for neural networks via nonlinear regression.
\newblock \emph{Technometrics}, 40\penalty0 (4):\penalty0 273--282, 1998.

\bibitem[Glorot and Bengio(2010)]{Glorot2010}
Xavier Glorot and Yoshua Bengio.
\newblock {Understanding the difficulty of training deep feedforward neural
  networks}.
\newblock In \emph{Proceedings of the Thirteenth International Conference on
  Artificial Intelligence and Statistics}, pages 249--256, 2010.

\bibitem[Goodfellow et~al.(2016)Goodfellow, Bengio, and
  Courville]{Goodfellow2016}
Ian Goodfellow, Yoshua Bengio, and Aaron Courville.
\newblock \emph{{Deep Learning}}, volume~1.
\newblock MIT Press Cambridge, 2016.

\bibitem[Hornik et~al.(1989)Hornik, Stinchcombe, and
  White]{hornik1989multilayer}
Kurt Hornik, Maxwell Stinchcombe, and Halbert White.
\newblock Multilayer feedforward networks are universal approximators.
\newblock \emph{Neural networks}, 2\penalty0 (5):\penalty0 359--366, 1989.

\bibitem[Hwang and Ding(1997)]{hwang1997prediction}
JT~Gene Hwang and A~Adam Ding.
\newblock Prediction intervals for artificial neural networks.
\newblock \emph{Journal of the American Statistical Association}, 92\penalty0
  (438):\penalty0 748--757, 1997.

\bibitem[Khosravi et~al.(2010)Khosravi, Nahavandi, Creighton, and
  Atiya]{khosravi2010lower}
Abbas Khosravi, Saeid Nahavandi, Doug Creighton, and Amir~F Atiya.
\newblock Lower upper bound estimation method for construction of neural
  network-based prediction intervals.
\newblock \emph{IEEE transactions on neural networks}, 22\penalty0
  (3):\penalty0 337--346, 2010.

\bibitem[Kingma and Ba(2014)]{Kingma2014}
Diederik~P Kingma and Jimmy Ba.
\newblock {Adam: {A} method for stochastic optimization}.
\newblock \emph{arXiv preprint arXiv:1412.6980}, 2014.

\bibitem[Kivaranovic et~al.(2019)Kivaranovic, Johnson, and
  Leeb]{kivaranovic2019adaptive}
Danijel Kivaranovic, Kory~D Johnson, and Hannes Leeb.
\newblock Adaptive, distribution-free prediction intervals for deep neural
  networks.
\newblock \emph{arXiv preprint arXiv:1905.10634}, 2019.

\bibitem[LeCun et~al.(2015)LeCun, Bengio, and Hinton]{LeCun2015}
Yann LeCun, Yoshua Bengio, and Geoffrey Hinton.
\newblock {Deep learning}.
\newblock \emph{Nature}, 521\penalty0 (7553):\penalty0 436, 2015.
\newblock ISSN 1476-4687.
\newblock \doi{10.1038/nature14539}.

\bibitem[Lei et~al.(2018)Lei, G’Sell, Rinaldo, Tibshirani, and
  Wasserman]{lei2018distribution}
Jing Lei, Max G’Sell, Alessandro Rinaldo, Ryan~J Tibshirani, and Larry
  Wasserman.
\newblock Distribution-free predictive inference for regression.
\newblock \emph{Journal of the American Statistical Association}, 113\penalty0
  (523):\penalty0 1094--1111, 2018.

\bibitem[Vovk et~al.(2005)Vovk, Gammerman, and Shafer]{vovk2005algorithmic}
Vladimir Vovk, Alex Gammerman, and Glenn Shafer.
\newblock \emph{Algorithmic learning in a random world}.
\newblock Springer Science \& Business Media, 2005.

\bibitem[Zhang et~al.(2019)Zhang, Zimmerman, Nettleton, and
  Nordman]{zhang2019random}
Haozhe Zhang, Joshua Zimmerman, Dan Nettleton, and Daniel~J Nordman.
\newblock Random forest prediction intervals.
\newblock \emph{The American Statistician}, \penalty0 (just-accepted):\penalty0
  1--20, 2019.

\end{thebibliography}


\end{document}